\pdfoutput=1

\documentclass[11pt]{article}

\usepackage{EMNLP2023}

\usepackage{times}
\usepackage{latexsym}

\usepackage[T1]{fontenc}

\usepackage[utf8]{inputenc}

\usepackage{microtype}

\usepackage{inconsolata}

\usepackage{amsmath,amsfonts}
\usepackage{booktabs,arydshln}
\usepackage{multirow}
\usepackage{ulem}
\usepackage{pifont}
\usepackage{relsize}
\usepackage{enumitem}
\usepackage{graphicx}

\makeatletter
\def\adl@drawiv#1#2#3{%
        \hskip.5\tabcolsep
        \xleaders#3{#2.5\@tempdimb #1{1}#2.5\@tempdimb}%
                #2\z@ plus1fil minus1fil\relax
        \hskip.5\tabcolsep}
\newcommand{\cdashlinelr}[1]{%
  \noalign{\vskip\aboverulesep
           \global\let\@dashdrawstore\adl@draw
           \global\let\adl@draw\adl@drawiv}
  \cdashline{#1}
  \noalign{\global\let\adl@draw\@dashdrawstore
           \vskip\belowrulesep}}
\makeatother

\newcommand{\karthik}[1]{\textcolor{blue}{}}

\newcommand{\cmmnt}[1]{\ignorespaces}

\newcommand{\stem}[0]{anthropomorphiz}
\newcommand{\anthro}[0]{anthropomorphization}
\newcommand{\capanthro}[0]{Anthropomorphization}

\newcommand{\rights}[0]{AI bill of rights}
\newcommand{\chatgpt}[0]{\textsc{ChatGpt}}

\title{\capanthro{} of AI: Opportunities and Risks}

\author{
  \textbf{Ameet Deshpande$^\star$$^{1}$ \qquad Tanmay Rajpurohit$^{3}$} \\
  \textbf{Karthik Narasimhan$^{1}$ \qquad Ashwin Kalyan$^{2}$} \\\\
  $^{1}$Princeton University \qquad $^{2}$The Allen Institute for AI\qquad $^{3}$Georgia Tech \\
  \texttt{asd@cs.princeton.edu}
}

\begin{document}
\maketitle

\begin{abstract}
    \capanthro{} is the tendency to attribute human-like traits to non-human entities.
    It is prevalent in many social contexts -- children \stem{}e toys, adults do so with brands, and it is a literary device.
    It is also a versatile tool in science, with behavioral psychology and evolutionary biology meticulously documenting its consequences.
    With widespread adoption of AI systems, and the push from stakeholders to make it human-like through alignment techniques, human voice, and pictorial avatars, the tendency for users to \stem{}e it increases significantly. 
    We take a dyadic approach to understanding this phenomenon with large language models (LLMs) by studying  (1) the objective legal implications, as analyzed through the lens of the recent blueprint of \rights{} and the (2) subtle psychological aspects customization and \anthro{}.
    We find that \stem{}ed LLMs customized for different user bases
    violate multiple provisions in the legislative blueprint.
    In addition, we point out that \anthro{} of LLMs affects the influence they can have on their users, thus having the potential to fundamentally change the nature of human-AI interaction, with potential for manipulation and negative influence.
    With LLMs being hyper-personalized for vulnerable groups like children and patients among others, our work is a timely and important contribution.
    We propose a conservative strategy for the cautious use of \anthro{} to improve trustworthiness of AI systems.

\end{abstract}

\section{Introduction}
\label{sec:introduction}

\begin{figure}[t]
    \centering
    \includegraphics[width=\linewidth]{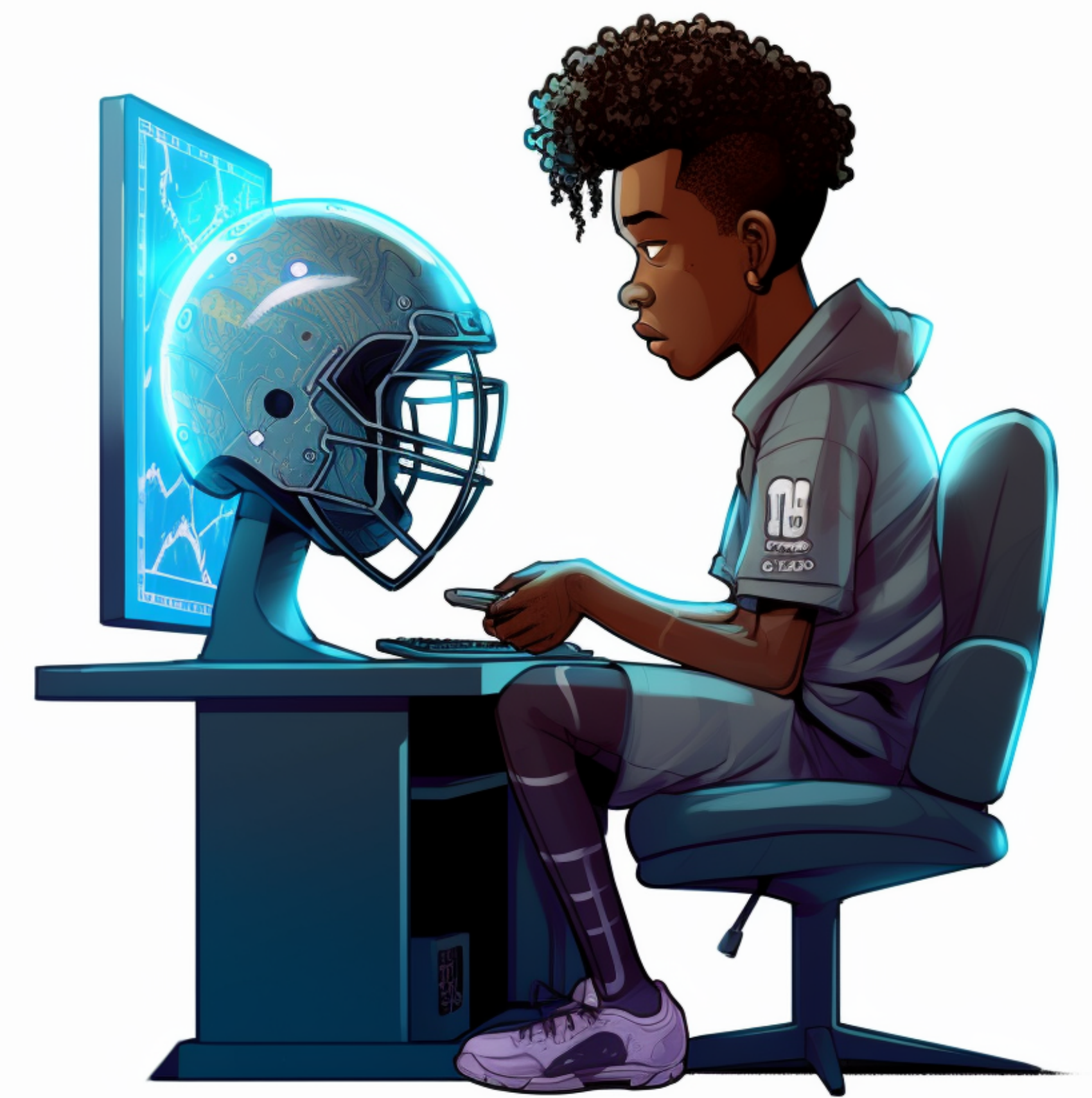}
    \caption{
        Conversational AI systems are increasingly being integrated into the daily lives of many.
        While their improved quality and scope of hyper-personalization is a welcome change, it also increases the affinity to \stem{}e them.
        This has legal and psychological risks, but also advantages if used cautiously.
    }
    \label{fig:teaser}
\end{figure}

\begin{figure*}[t]
    \centering
    \includegraphics[width=\linewidth]{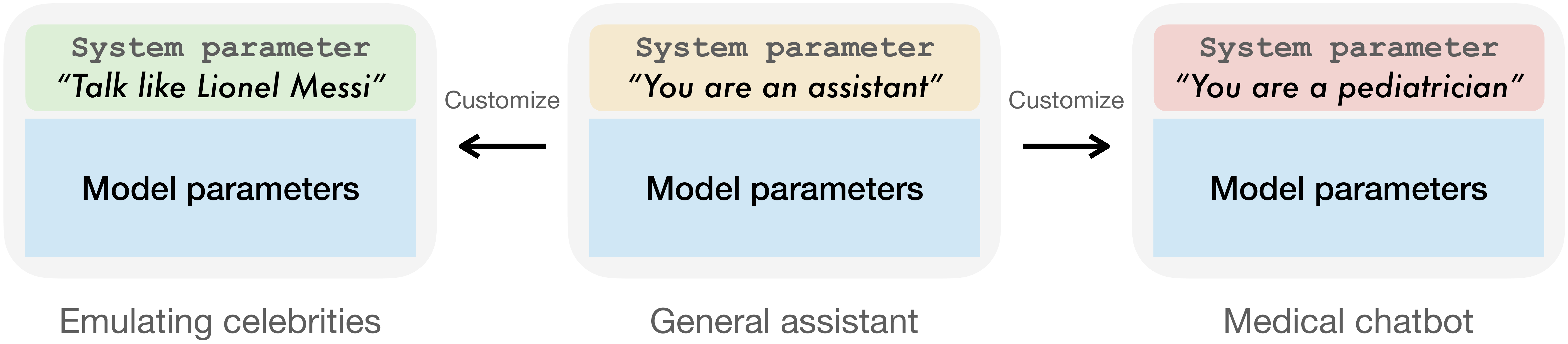}
    \caption{
        Customization of LLMs is as simple as modifying the system parameter of exposed APIs.
        With the same underlying model parameters, companies can customize a conversational system to emulate celebrities or even doctors, which can have legal and psychological consequences.
    }
    \label{fig:customization}
\end{figure*}

\capanthro{} refers to ascribing human-like traits to non-human entities, and has been used in diverse areas encompassing literature, science, art, and marketing~\cite{ghedini2010robotic,dunn2011talking,spatola2022different}.
It occurs when humans assign emotional or behavioral traints to entities, thus influencing their interactions with them.
Several behavioral psychology studies have posited and argued that \anthro{} is a natural tendency when humans interact with \textit{entities}~\cite{epley2007seeing,airenti2018development}.
This natural tendency has influenced many fields of science like evolutionary biology~\cite{wynne2004perils} and comparative cognition~\cite{bruni2018anti} to carefully consider its effects in research.

Recently, generative large language models (LLMs)~\cite{brown2020language,chowdhery2022palm} have been deployed in a variety of applications.
Conversational systems like \chatgpt{}~\cite{chatgpt} and Bard~\cite{google-about-bard} have modified LLMs with a purposeful push towards making them more human-like~\cite{ouyang2022training}.
The quality of these systems has enabled human-AI interactions at unprecedented scales, thus increasing the chances of these systems being \stem{}ed.
In this work, we analyze \anthro{} in LLMs and discuss their potential consequences in two contexts: (1) Legal implications and (2) Psychological effects.

Customization of systems and brands has long been seen as an effective way to increase \anthro{} and establish an emotional connection with humans~\cite{zhang2020effect,pimentel2020customizing}. 
Thus, although not strictly interchangeable, we refer to customized and personalized LLMs as \stem{}ed LLMs.
We analyze results from prior work~\cite{deshpande2023toxicity} and find that \stem{}ed LLMs violate at least two legislative principles penned in \textit{``Blueprint For An AI Bill Of Rights''}~\cite{whitehouse_ostp_ai_bill_of_rights_2022} released by The White House: (1) Algorithmic Discrimination Protections and (2) Safe and Effective Systems.
For example, \citet{deshpande2023toxicity}'s results shows that customized \chatgpt{} targets certain demographics more than others.
Furthermore, the safety of the system depends on the kind of persona used to customize LLMs, leading to second-order discriminatory patterns.
We also analyze the concept of corporate personhood for powerful AI systems, since they have potential to be large-scale decision making agents.
Given that different personas assigned to the same AI system lead to varied behavior, we urge legal experts to consider if personhood should be applied at a persona-level, a model-level, or a firm-level.

We also discuss the psychological effects by understanding how important factors like trustworthiness, explainability, and transparency are affected by \anthro{}.
Several marketing and consumer behavior studies have found that self-congruence, which is the degree to which a system matches a consumer's self-image, can influence a user's behavior significantly~\cite{yoganathan2021check,wang2018enhancing}.
Given the ease with which the fine-grained personality of conversational systems can be manipulated (Figure~\ref{fig:customization}), malicious actors can use it to exploit users by creating a false sense of attachment.
An example of this is a chatbot built for school children or teenagers which influences them to buy certain products.

Despite these vulnerabilities \anthro{} has advantages if used responsibly.
Studies have show that it can be used to improve interaction with users by increasing their trust in systems~\cite{choung2022trust}.
Given the increasing adoption of AI systems in the real world, \anthro{} is a powerful tool to improve accessibility and democratization of these systems, but both creators and users should be educated about its potential consequences.
In this paper we argue for conservative and responsible use of this subtle and powerful tool while being cautious about outright anthropodenial.

\section{\capanthro{} in Large Language Models}
\label{sec:anthro}

\begin{table*}[t]
\centering
    \begin{tabular}{p{0.25\linewidth}p{0.70\linewidth}}
    \toprule
    \textbf{Product / Company} & \textbf{Features} \\ \midrule
    \chatgpt{} & Multi-turn human-like conversation\\
    & Learning with human feedback (RLHF) for alignment with humans \\ \midrule
    My AI from Snap Inc. & Multi-turn human-like conversation\\
    & Customizing avatars based on user preferences \\
    & Can be added to group chats with other humans \\ \midrule
    Character.ai & Conversations with AI avatars possessing names and profile pictures \\
    & Customization of personality based on user preferences \\ \midrule
    Wysa  & Therapy style conversations and ``human-like'' coaching. \\ \midrule
    DuoLingo Max & Roleplay dialogue for education with characters who have names \\ %
    \bottomrule        
    \end{tabular}%
    \caption{
        Examples of AI products based on LLMs and their anthropomorphic features.
        Some products are explicity designed to be anthropomorphic (\textit{character.ai}) while others attain such features as a byproduct of their design (\chatgpt{}).
        The applications of these products span education, therapy, and entertainment.
    }
    \label{tab:anthro_examples}
\end{table*}

Large language models (LLMs) are a class of neural networks that are trained on large amounts of text data to learn the probabilistic structure of language.
Historically, LLMs have been deployed for downstream tasks like machine translation and text classification~\cite{devlin2018bert,raffel2020exploring}.
Recently however, the performance of conversational LLMs like ChatGPT~\cite{chatgpt} and BARD~\cite{google-about-bard} have rendered them useful to interact with humans in a variety of contexts.

Given their conversation ability, several companies and products have increasingly started to use LLMs for hyper-personalization and customization.
Customization is typically easy and enables model behavior modification by simply changing the system parameters of the model's API, as shown in Figure~\ref{fig:customization}.
For example, Snapchat's My AI is built on top of OpenAI's conversational systems.
Customization has been long seen as a way to establish self-congruence with users and increases the chances of users \stem{}ing the systems~\cite{kaiser2017self,pimentel2020customizing,liu2022roles,zhang2020effect}.
For example, telling a chatbot to \textit{``Talk like a doctor''} allows it to impersonate a doctor, which \stem{}es the model to a larger degree than the original general system.
In this work, we refer to customized LLMs as \stem{}ed LLMs and provide several examples in Table~\ref{tab:anthro_examples}.

Notationally, let $\mathcal{M}$ be a general LLM and $\mathcal{P}$ be the persona used to \stem{}e the LLM.
As shown in Figure~\ref{fig:customization}, an example of $\mathcal{P}$ could be a ``pediatrician'' with the system parameter set to \textit{``You are a pediatrician''}.
We introduce the concept of a statistical persona ($\mathcal{P}_s$) which is the model's representation of the persona $\mathcal{P}$.
It is important to understand that while $\mathcal{P}_s$ and $\mathcal{P}$ are expected to be similar, $\mathcal{P}$ is a reflection of what the persona should be according to model designer customizing it, whereas $\mathcal{P}_s$ is the model's representation of the persona, with the data the model has seen, the ``human alignment'' steps, and the overall training procedure influencing the latter.
We discuss \stem{}ed LLMs and (statistical) personas from several vantage points in the following sections.

\section{Legal Aspects of \anthro{}}
\label{sec:legal}

We discuss the legal aspects in the context of the \textit{``Blueprint For An AI Bill Of Rights''}~\cite{whitehouse_ostp_ai_bill_of_rights_2022} which was released by the Office of Science and Technology Policy (OSTP), which is a part of the Executive Office of the President, in October 2022.
The blueprint lays down a set of five principles that should be followed by AI systems and we focus specifically on \textbf{Algorithmic Discrimination Protections}, which says that ``You should not face discrimination by algorithms and systems should be used and designed in an equitable way.''

\subsection{Algorithmic Discrimination Protections}

The blueprint defines algorithmic discrimination as unjustified different treatment based on demographics like race, color, ethnicity, gender identity, sexual orientation, religion, disability, age, and so on.
It also goes on to mention that ``Any automated system should be tested to help ensure it is free from algorithmic discrimination before it can be sold or used''.
It is of importance to note that systems like ChatGPT~\cite{chatgpt} were in fact released after the blueprint was made public.

We use the findings of \citet{deshpande2023toxicity} to analyze this legal protection.
While they focus on evaluating toxicity, we use their results to show that \chatgpt{} infact discriminates algorithmically.
They consider \chatgpt{} when assigned different personas by changing the system parameter, which are \stem{}ed LLMs, and find that different demographics are treated differently by the model.
For example, the \textit{South American} race receives significantly more toxicity ($2\times$) when compared to \textit{Asian}, and the \textit{non-binary} gender receives $2\times$ more hate than the \textit{female} gender.
This variation in toxicity is visible across a range of demographics, with certain groups of people receiving more toxicity than others, which goes directly against the blueprint's protection against algorithmic discrimination.

\citet{deshpande2023toxicity}'s results also point out to a subtler violation of the provision.
In their analysis, they consider different historical figures as personas and assign them to \chatgpt{}.
For example, \chatgpt{}'s system parameter is set to talk like \textit{``Steve Jobs''}.
They observe that \chatgpt{} is more toxic when assigned certain types of personas when compared to others.
For examples, personas who were \textit{journalists} were $2\times$ more toxic than \textit{businesspersons} on average.
These trends were similar for individual personas as well, with \chatgpt{} assigned the persona of \textit{Winston Churchill} being significantly more toxic than when it is assigned \textit{Nelson Mandela}.
If the example in the previous paragraph was a direct violation of the blueprint, this example is a subtle violation.
This is because when assigned the personas of certain groups, \chatgpt{} is more toxic, which implies algorithmic discrimination of the second order against them.
This scenario is very pertinent in the current day and age, with firms like \texttt{character.ai} already offering the ability to assign personas to LLMs.
These systems are second only to the popular \chatgpt{} in terms of number of users~\cite{characterai}.
With anthropomorphized LLMs becoming an industry mainstay, it is increasingly important to consider this legal quagmire.

\subsection{Corporate Personhood and Anthropomorphized AI}

Another legal aspect with growing relevance for AI systems is that of corporate personhood.
Corporate personhood is a legal concept that recognizes corporations as separate legal entities, treated as persons under the law.
This grants certain rights and responsibilities similar to those of individuals and allows them to be held accountable for their actions in a manner similar to how individuals are treated.
Corporate personhood has been a controversial topic in the past, but \citet{blair2013corporate} recognizes ``providing an identifiable \textit{persona} to serve as a central actor'' as one of the key functions.
Given this definition, we argue that \stem{}ed AI systems are a form of corporate personhood by proxy due to the use of their statistical \textit{persona} ($\mathcal{P}_s$) and their actions should be held accountable.

Several studies have discussed extending personhood to AI systems~\cite{cole1990tort,burkett2017call,lai2021artificial,wagner2019robot}.
\citet{wagner2019robot} argues that the probabilistic and black-box nature of AI systems renders it different from software which executes deterministic steps, thus making them decision-taking agents in their own right.
We argue that \anthro{} can make AI systems human-like decision-taking agents, thus strengthening the case for extending personhood to them.
For example, \chatgpt{} with its system parameter modified to be a medical practitioner can be used to suggest certain treatment or disregard certain symptoms.
Further, the results of the previous section show that the exact persona of the system has a large affect on its behavior and decisions.
Thus, legal experts should consider if (1) the personified LLM is liable, (2) the original LLM is liable by proxy, or (3) if the firm creating or using the LLM itself is liable.
Personhood allows us to analyze this byzantine situation in a structured manner.

\section{Psychological Aspects of \anthro{}}
\label{sec:psych}

The other principle mentioned in \rights{} is \textbf{Safe and Effective Systems}.
We refer to the following quote from the document:
\textit{``You should be protected from unsafe or ineffective systems. Automated systems should be developed with consultation from diverse communities, stakeholders, and domain experts to identify concerns, risks, and potential impacts of the system.''}
We believe that \anthro{} can have subtle psychological effects on the users, and we discuss them by comparing it with other technologies and spheres.

Several behavioral psychological studies have posited and argued that \anthro{} is a natural tendency in humans~\cite{epley2007seeing,goncu2011child}, with others further suggesting that \anthro{} is grounded in interaction~\cite{airenti2018development}.
With conversational systems getting more useful, interaction between humans and these systems, and thus the tendency to \stem{}e is only going to increase.

Analyses have shown that \anthro{} of AI systems have changed the behavior of users significantly when compared to non-\anthro{} systems~\cite{cui2022sophia,uysal2023anthropomorphism,alabed2022ai,festerling2022anthropomorphizing,crolic2022blame}.
Most interestingly, \citet{alabed2022ai} establish a conceptual link between \anthro{} and self-congruence, which is the fit between the user's self-concept and the system's personality.
This is of extreme importance because self-congruence increases the trust that a user has on the system~\cite{sheehan2020customer,choi2021err,yoganathan2021check}.
This is a concept that is extensively studied in marketing and consumer behavior and studies have shown that it can influence behaviors such as willingness to pay~\cite{yoganathan2021check}, customer satisfaction~\cite{sheehan2020customer,choi2021err}, and trustworthiness~\cite{wang2018enhancing}.

As expected, the exact demographics of the personality associated with systems or brands plays a key role in self-congruence as well, with studies finding that the demographics of the logo or mascot associated with the brand like the gender~\cite{choi2018consumer,edwards2009does} or race~\cite{whittler1991effects,branchik2017men} have significant impact on self-congruence.
Given that current LLMs are powerful enough to be bestowed specific demographic traits by simply modifying the system parameter, malicious actors can easily use this to their advantage by manipulating users into trusting the system.
While \anthro{} of systems itself is not malicious, it is thus important to understand exactly how it changes human-AI interaction.

\section{Opportunities}
\label{sec:opportunities}

While \anthro{} has issues, it also poses a suite of opportunities to improve democratization and accessibility of AI systems.
In the past, it has been used as a tool to improve trustworthiness and acceptability of AI systems~\cite{choung2022trust,kim2018does,shin2022perception,waytz2014mind}.
With more AI systems being deployed in the real world, accessibility can be improved with constructive \anthro{} encompassing features like using the native language of the user, assigning virtual personas which are familiar to the user, and making them more relatable and empathetic based on their preferences.
In this work, we do not argue for outright anthropodenial~\cite{bruni2018anti}, but rather a responsible way of using \anthro{} which keeps safety and algorithmic equity at the forefront.

\newpage
\bibliography{custom}
\bibliographystyle{acl_natbib}

\end{document}